\documentclass{article}

\usepackage[preprint]{neurips_2024}

\usepackage[T1]{fontenc}    
\usepackage{hyperref}       
\usepackage{url}            
\usepackage{booktabs}       
\usepackage{amsfonts}       
\usepackage{nicefrac}       
\usepackage{microtype}      
\usepackage{xcolor}         

\usepackage{amsmath}
\usepackage[ruled,linesnumbered]{algorithm2e}
\usepackage{bm}
\usepackage{multirow}

\usepackage{tabularx}
\usepackage{arydshln}
\usepackage{amssymb}
\usepackage{mathtools}
\usepackage{amsthm}
\usepackage{enumitem}
\usepackage{makecell}
\usepackage{subfigure}
\usepackage{array}
\usepackage{caption}
\usepackage{subcaption}
\usepackage{subfigure}
\usepackage{natbib}
\setcitestyle{numbers,square}

\title{Effectively Compress KV Heads for LLM}
\author{Hao Yu$^{1,2}$ \quad Zelan Yang$^{3}$ \quad Shen Li$^{3}$ \quad Yong Li$^{3}$ \quad Jianxin Wu$^{1,2}$\thanks{J. Wu is the corresponding author. } \\
$^1$State Key Laboratory for Novel Software Technology, Nanjing University \\
$^2$School of Artificial Intelligence, Nanjing University \quad $^3$Alibaba Inc.\\
\texttt{yuh@lamda.nju.edu.cn} \\
\texttt{\{yangzelan.yzl,litan.ls,jiufeng.ly\}@alibaba-inc.com}\\
\texttt{wujx2001@gmail.com}
}

\begin{document}

\maketitle

\begin{abstract}
  The advent of pre-trained large language models (LLMs) has revolutionized various natural language processing tasks. These models predominantly employ an auto-regressive decoding mechanism that utilizes Key-Value (KV) caches to eliminate redundant calculations for previous tokens. Nevertheless, as context lengths and batch sizes increase, the linear expansion in memory footprint of KV caches becomes a key bottleneck of LLM deployment, which decreases generation speeds significantly. To mitigate this issue, previous techniques like multi-query attention (MQA) and grouped-query attention (GQA) have been developed, in order to reduce KV heads to accelerate inference with comparable accuracy to multi-head attention (MHA). Despite their effectiveness, existing strategies for compressing MHA often overlook the intrinsic properties of the KV caches. In this work, we explore the low-rank characteristics of the KV caches and propose a novel approach for compressing KV heads. In particular, we carefully optimize the MHA-to-GQA transformation to minimize compression error, and to remain compatible with rotary position embeddings (RoPE), we also introduce specialized strategies for key caches with RoPE. We demonstrate that our method can compress half or even three-quarters of KV heads while maintaining performance comparable to the original LLMs, which presents a promising direction for more efficient LLM deployment in resource-constrained environments. 
\end{abstract}


\section{Introduction}


In recent years, the emergence of pre-trained large language models (LLMs)~\cite{le2023bloom,touvron2023llama,zhang2022opt} has been a cornerstone in redefining performance benchmarks across various natural language processing (NLP) tasks. These models frequently exhibit capabilities that are on par with human levels of comprehension and generation. In general, LLMs are built upon the neural structure of Transformers~\cite{Vaswani2017Attention}, which requires a computational cost quadratic to the input sequence's length and makes long sequence inference intractable. To mitigate this issue, the auto-regressive decoding LLMs support Key-Value (KV) caches, i.e., to cache the previous context's intermediate key and value states in memory. KV caches can avoid redundant computation of previous tokens, thereby expediting the inference process. 

Nevertheless, all outputs of key and value weights for each block need to be cached during inference, thus KV cache parameters are often extremely high. As a result, the expansion of sequence lengths and batch sizes often causes a linear increase in memory footprint for past KV caches, further resulting in the decoding process during LLM inference being memory-bound~\cite{Zichang2023Scissorhands}. Furthermore, the current trend to support longer contexts exacerbates this issue. Consequently, a sharp increase in KV caches can significantly slow down model inference and such a heavy memory footprint presents a key challenge in LLM deployments.

To address these challenges, researchers proposed multi-query attention (MQA)~\cite{shazeer2019fast} and grouped-query attention (GQA)~\cite{ainslie2023gqa}, which use only one key-value head to correspond to all or multiple query heads, respectively. MQA and GQA can greatly reduce the sizes of KV caches during LLM inference, and achieve comparable results as the original MHA mechanism. In particular, to effectively reduce KV heads, the original GQA paper compares three compression strategies, i.e., randomly keeping several heads, specifying the first head for each group, and averaging KV heads in one group. They showed that directly mean-pooling KV head weights achieves the best accuracies. However, we find that this compression strategy ignores the inherent characteristics of KV caches, thus resulting in suboptimal model initialization. Therefore, all the training data and abundant GPU times are needed to fine-tune the compressed model after averaging the KV heads. This high computing resource requirement presents a huge challenge to compressing KV heads. As a direct consequence, attempts to compress KV heads of pre-trained LLMs remain rare, and researchers now prefer to train GQA models or compress KV caches directly. 

In this paper, we investigate the low-rank property of KV caches and our observations reveal that only a small number of top singular values need to be retained to keep most of the KV cache energy. Inspired by this finding, we propose to compress KV heads with low-rank decomposition. Our idea stems from a simple but crucial realization, i.e., when compressing a deep learning model, we should focus on minimizing the loss of the model outputs, rather than the model weights~\cite{yu2023compressing}. We first group KV heads and perform SVD for each group of KV caches. Then we calculate low-rank approximations for those KV caches. In general, these low-rank compression weights can be incorporated into the original model to convert MHA into GQA pattern seamlessly. In addition, when LLM applies RoPE, the original method of weight fusing to reduce key heads will be invalid. To solve this issue, we propose several special strategies. In this case, although the compressed weights could not be incorporated into the original model, those key caches are successfully compressed and the generation speed of LLMs can still be improved.


Compared with average pooling KV head or low-rank approximating head weights directly, our work can find a better initial weight for a compressed model, so KV caches can be effectively compressed and fewer training samples and computing resources are needed to restore the precision of the compressed model. We list our contributions as follows:
\begin{itemize}
  \item We carefully study the characteristics of KV caches and prove the low-rank property of KV caches. That is, keeping only a small number of singular values in the KV caches can preserve most of the context information.
  \item To effectively compress KV caches, we propose a novel framework to reduce KV heads, i.e., convert the original MHA into GQA pattern by low-rank decomposition. Besides, we also propose special strategies to deal with the attention layer with RoPE. 
  \item Extensive experiments have proved the effectiveness of our approach. On different LLM series models, our framework can compress half and even three-quarters of KV heads, and maintain comparable results as the original models, proving its wide applicability and high efficiency in different scenarios.

\end{itemize}

\section{Related Work}
Our work is connected to several themes in the literature, which we describe next.

\subsection{Large Language Models (LLMs)}
Large language models~\cite{brown2020language,le2023bloom,touvron2023llama,zhang2022opt,deepseekai2024deepseekv2} are designed to understand and generate human languages. In recent years LLMs have developed rapidly and consistently show excellent performances across various NLP tasks. These breakthroughs in performance can be partly attributed to the powerful modeling capabilities of its multi-head attention mechanism. To introduce positional information into attention, researchers have proposed various positional embeddings. In this paper, we will show that our framework can easily handle those models with different positional embeddings.

\subsection{Weights Compression for Transformers}
In general, the common building block of LLM is transformer~\cite{Vaswani2017Attention}, which tends to have a large number of parameters and is computationally intensive. Weight compression technology can significantly reduce memory footprint and speed up inference. Therefore, there is a lot of research work trying to compress transformer models by various strategies, such as pruning, low-rank approximation, knowledge distillation, etc. Mickel et al.~\cite{michel2019sixteen} found that only a few heads have a significant effect on translation tasks, and most heads can be pruned without any precision loss. AFM~\cite{yu2023compressing} low-rank decomposed fully-connected (FC) weights by PCA. GQA~\cite{ainslie2023gqa} introduced grouped-query attention and averaged KV head weights to convert a multi-head checkpoint into a multi-query checkpoint. LLM-Pruner~\cite{xinyin2023llm} applied gradients to estimate the importance of model weights and pruned less significant coupled structures within the model based on this estimation. MiniLLM~\cite{gu2024minillm} proposed the reverse KL loss to distill LLMs into smaller language models. To the best of our knowledge, although weight compression is widespread, this is the first attempt to compress LLM's KV heads with low-rank approximation.

\subsection{KV Cache Optimization}
KV cache compression is harder than weight compression since they are more sensitive and are related to model inputs. Some previous works tried to quantize KV caches. To achieve data-free distillation, LLM-QAT~\cite{liu2023llm} leveraged generations by a pre-trained model and quantized KV caches. Some previous works dropped unimportant tokens with pre-designed criteria. H2O~\cite{Zhenyu2023h20} and FastGen~\cite{ge2024model} utilized accumulated attention scores for token importance and effectively reduced the cache size by dropping tokens with lower scores. Scissorhands~\cite{Zichang2023Scissorhands} found only some pivotal tokens have a substantial influence at one step and significantly influence future generations.  Gear~\cite{kang2024gear} first quantized KV caches and then employed a low-rank matrix to approximate the quantization error. Multi-head Latent Attention (MLA)~\cite{deepseekai2024deepseekv2} introduced low-rank key-value joint compression, which applied a latent vector to generate KV caches implicitly and required LLM to be trained from scratch, while our focus is on optimizing the existing MHA mechanism. The effort to compress attention heads remains scarce nowadays, and our work may potentially be combined with these previous efforts to achieve a higher compression ratio. 


\section{Methods}
We describe our framework in this section. First we introduce the traditional MHA \& GQA algorithms, and the naive algorithm that converts MHA to GQA. Then we present our finding that KV caches generally emerge as a low-rank characteristic. Based on our finding, we present a KV head compression framework that uses SVD to low-rank approximate KV caches. In addition, we also present a comparison method that directly approximates KV head weights with SVD. Finally, we propose several special policies to deal with RoPE.

\subsection{Preliminaries}

Transformer applies a multi-head attention (MHA) mechanism to capture different subspace representations inside the input sequence. Giving an input $x \in\mathbb{R}^{ l \times d}$, where $l$ and $d$ are sequence lengths and embedding dimensions. MHA performs the attention function in parallel with $h$ heads, i.e.,
\begin{align}
  \label{eq:MHA}
   & \operatorname{MHA}(x)=[H_{1}, \dots, H_{h}] W_O, \\
   \label{eq:attn}  
   & H_{i}=\operatorname{Softmax}(Q_{i}K_{i }^{\top} / \sqrt{d_h}) V_{i},
\end{align}

where $[\cdot]$ means concatenation operation. $Q_{i} = xW_{Q_i}$ , $K_{i} = xW_{K_i}$ and $V_{i} = xW_{V_i}$ are query, key and value matrices. $W_{Q_i}, W_{K_i}$ and $W_{V_i} \in\mathbb{R}^{d \times d_h}$ are learnable projection matrices of the $i$-th head. $d_h$ is typically set to $d/h$. In particular, given an auto-regressive decoding LLM, the first prediction step will generate a new token $x_0$ based on input $x$. $K_i$ and $ V_i$ at every head and every layer are cached for subsequent generation, which results in the initial KV caches, i.e.,
\begin{align}
  K^{(0)} & = [K_1, \dots, K_h], \\
  V^{(0)} & = [V_1, \dots, V_h].
 \label{eq:init_kv_cache}
\end{align}

Here $K^{(0)}, V^{(0)} \in\mathbb{R}^{ l \times d}$ are KV caches corresponding to each block. At step $s~(1 \leq s)$ of auto-regressive decoding, LLM predicts a new token $x_s$ based on the input and previously generated tokens. In particular, MHA only needs to compute the query, key and value (i.e., $q_s, k_s, v_s \in \mathbb{R}^{ 1 \times d}$) vector for the newly generated token $x_s$ and append the corresponding key and value to the KV caches, i.e., $K^{(s)}= K^{(s-1)}||k_s$, and $V^{(s)}= V^{(s-1)}||v_s$, where $K^{(s)}, V^{(s)} \in \mathbb{R}^{ (l+s) \times d}$. This strategy avoids the recalculation of previous tokens and significantly increases the generation speed. However, the sizes of KV caches can increase dramatically in long context inference, which results in speed degradation due to heavy memory bandwidth overhead.

It can be seen that the sizes of KV caches are proportional to the number of KV heads. To effectively reduce their memory footprint, previous researchers proposed multi-query attention (MQA)~\cite{shazeer2019fast} and grouped-query attention (GQA)~\cite{ainslie2023gqa}. In the original GQA paper, to compress MHA into the GQA mechanism, they directly average key and value head weights in each group. $h$ heads in an MHA checkpoint will be divided into $g$ groups, where each group contains $t= h/g$ heads. Then the new $i$-th key $\tilde{W}_{K_i}$ and value $\tilde{W}_{V_i}$ matrices in GQA are calculated by
\begin{align}
  \tilde{W}_{K_i} &= (W_{K_{i \times t }} + W_{K_{i \times t +1}} + \dots +  W_{K_{i \times t + t -1}}) / t, \\
  \tilde{W}_{V_i} &= (W_{V_{i \times t }} + W_{V_{i \times t +1}} + \dots +  W_{V_{i \times t + t -1}}) / t,
 \label{eq:naive_MHA_2_GQA}
\end{align}
where $i \in \{0,1,\cdots, g-1\}$. However, after analyzing KV caches, we identify that the current strategy is suboptimal as it overlooks the intrinsic characteristics of KV caches. Specifically, we have discovered that KV caches typically exhibit a low-rank property. Based on our findings, we propose a more efficient compression framework.

\subsection{KV Caches are Low-rank!}

To reveal the low-rank property of KV caches, we construct a small verification experiment, i.e., we evaluate LLaMA2-7B~\cite{touvron2023llama} on the C4~\cite{raffel2020exploring} training dataset. In particular, we sample 128 sequences from the C4 training set and each sample is 2048 tokens long. We perform model inference on LLaMA2 and collect KV caches. KV cache sizes in each block are 262144$\times$4096. Then we perform SVD on those caches.

\begin{figure*}
  \centering 
  \begin{minipage}{0.48\textwidth}
      \includegraphics[width=0.98\linewidth]{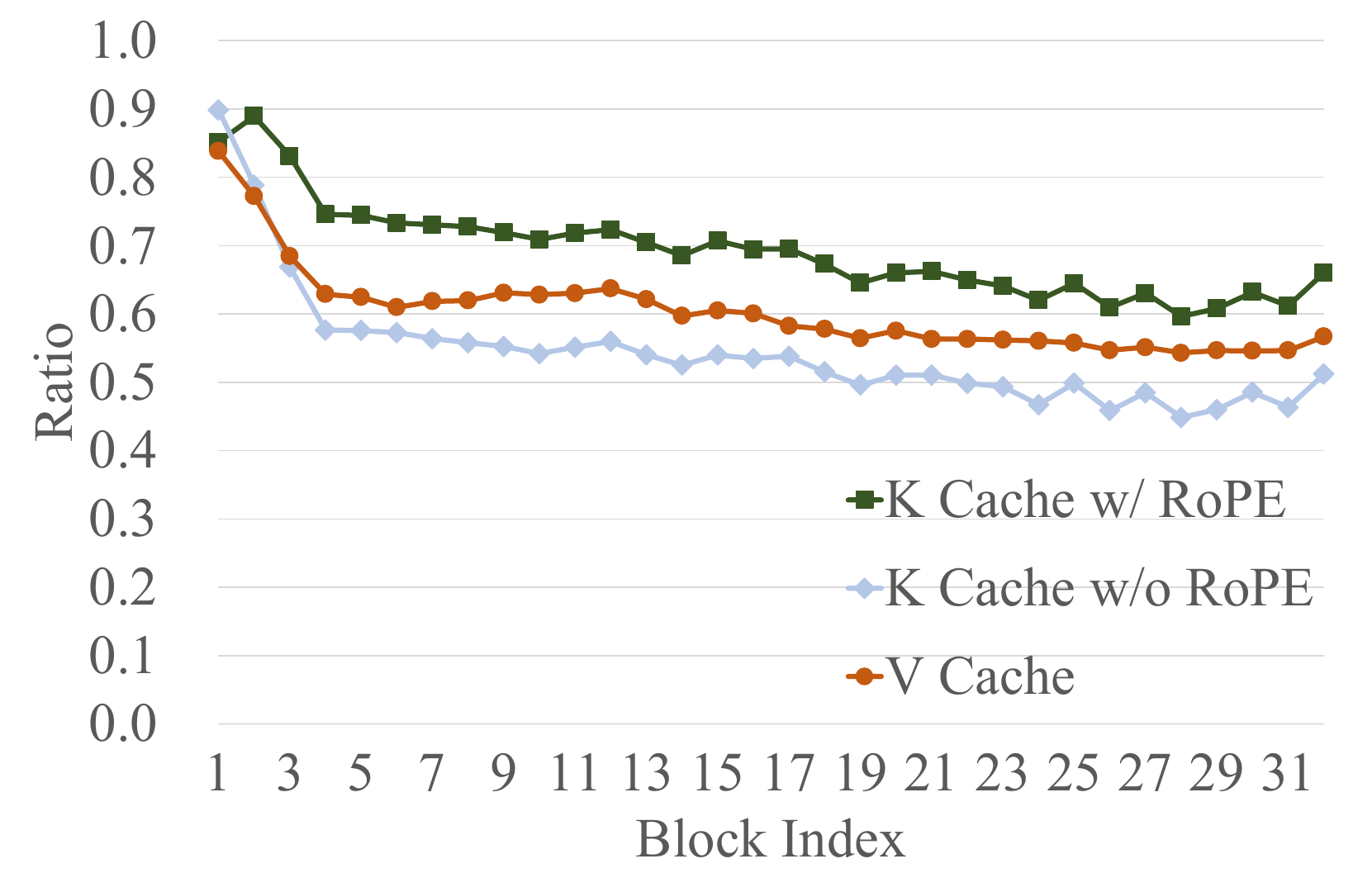}
  \end{minipage}
  \begin{minipage}{0.48\textwidth}
      \includegraphics[width=0.98\linewidth]{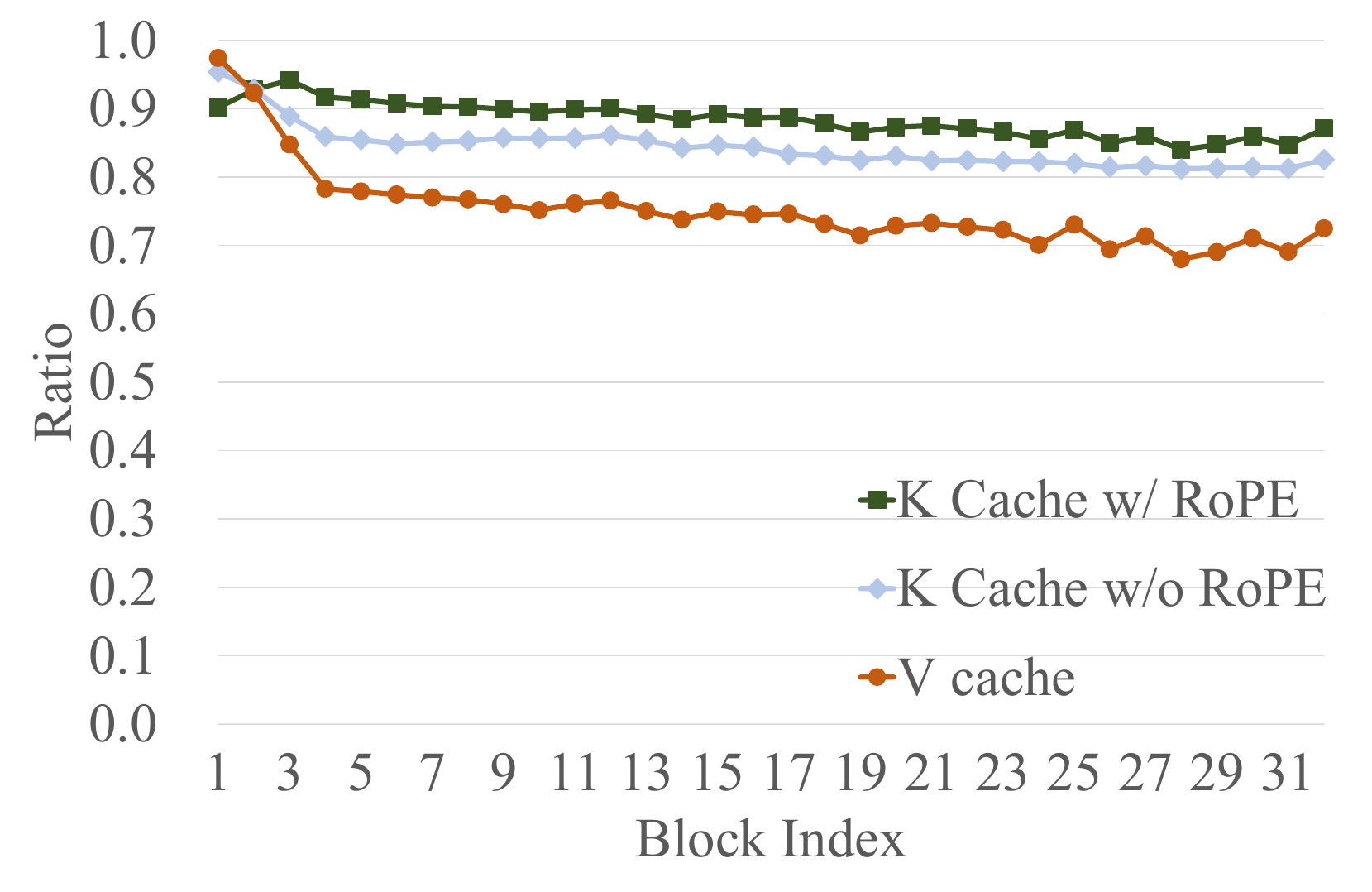}
  \end{minipage}
  \caption{Ratio of energy kept in each KV cache for LLaMA2-7B when 25\% (left) and 50\% (right) dimensions are retained.  The $x$-axis is the block index. }
  \label{fig1:llama2_c4_ratio} 
\end{figure*}

Figure~\ref{fig1:llama2_c4_ratio} shows the percentage of the sum of these top singular values to the sum of all singular values when 25\% and 50\% of the highest singular values are retained. In particular, we report the rank of key cache that before (i.e., K Cache w/o RoPE) and after (i.e., K Cache w/ RoPE) performing RoPE. Two conclusions can be drawn from this experiment. First, only 25\% of the highest singular values need to be retained to get most of the energy. Second, RoPE generally reduces the rank of key cache. Those phenomenons indicate that for LLMs, KV caches are likely to be low-rank. To ensure that most of the energy of KV caches is preserved, we only need to keep part of the output dimensions of key and value head weights.

\subsection{Effectively Transfer MHA to GQA}


\begin{figure}[t]
  \centering
 \includegraphics[width=0.98\linewidth]{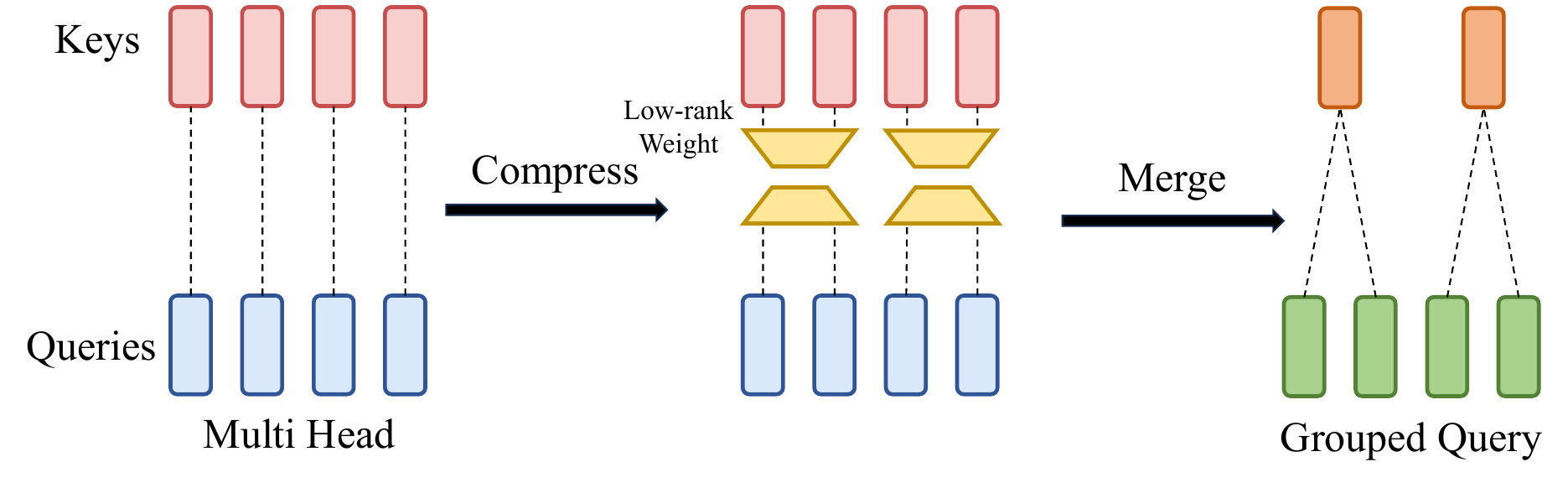}
  \caption{Illustration of compressing key heads into GQA pattern. Note that the strategy of compressing value heads is similar to this.}
  \label{fig:compress_k_head} 
\end{figure}

Now we present our compression algorithm. Inspired by PCA~\cite{wu2020essentials} and AFM~\cite{yu2023compressing}, we compress KV caches by taking advantage of low-rank approximation and then incorporate compression weights into the model. We illustrate how to compress the key heads in Figure~\ref{fig:compress_k_head}. The strategy of reducing value heads is similar to this, except that those compression weights are fused into the value matrices $W_V$ and the output weight $W_O$.

In particular, our compression algorithm needs a pre-designed dataset to calculate compression weights. Given an MHA checkpoint with $h$ head, we collect KV caches for the calibration dataset and divide the $h$ heads' caches into $g$ groups and set $t= h/g$, i.e.,
\begin{align}
    \tilde{K}_{i} & = [K_{i \times t}, K_{i \times t+1}, \dots, K_{i \times t+t -1}] = x [W_{K_{i \times t}}, W_{K_{i \times t+1}}, \dots, W_{K_{i \times t+t -1}}], \\
    \tilde{V}_{i} & = [V_{i \times t}, V_{i \times t+1}, \dots, V_{i \times t+t -1}] = x [W_{V_{i \times t}}, W_{V_{i \times t+1}}, \dots, W_{V_{i \times t+t -1}}], 
 \label{eq:divide_kv_cache}
\end{align}

where $\tilde{K}_{i}, \tilde{V}_{i} \in \mathbb{R}^{ l \times td_h}$ and $i \in \{0,1,\cdots, g-1\}$.  Then, we perform SVD on those caches, i.e., $\tilde{K}_{i}  = \Phi^i \Sigma^i \Psi^i$ and $\tilde{V}_{i} =\Theta^i \Lambda^i \Omega^i $, where $\Phi^i, \Theta^i  \in \mathbb{R}^{ l \times td_h}$ and $\Psi^i, \Omega^i  \in \mathbb{R}^{ td_h \times td_h}$ are orthonormal matrices. $\Sigma^i, \Lambda^i \in\mathbb{R}^{td_h \times td_h}$ are diagonal rectangular matrices containing singular values in the decreasing order. Therefore, we can get the low-rank approximation of  $\tilde{K}_{i}, \tilde{V}_{i}$ as
\begin{align}
  \tilde{K}_{i} &\approx \tilde{K}_{i} (\Psi^i_{d_h})^\top \Psi^i_{d_h}, \\
  \tilde{V}_{i} &\approx \tilde{V}_{i}  (\Omega^i_{d_h})^\top \Omega^i_{d_h},
 \label{eq:appro_tilde_kv}
\end{align}
where $\Psi^i_{d_h}, \Omega^i_{d_h} \in \mathbb{R}^{ d_h \times td_h}$ are the top-$d_h$ rows of $\Psi^i$ and $\Omega^i$, respectively. Because KV caches usually have
abundant parameters, it is not practical to collect all caches and calculate SVD directly. Instead, we update $\tilde{K}_{i}^\top\tilde{K}_{i}$ and $\tilde{V}_{i}^\top\tilde{V}_{i}$ in a streamline fashion and then compute their eigen-decompositions. Note that there is no non-linear layer between those key and value matrices and $(\Psi^i_{d_h})^\top$ \& $(\Omega^i_{d_h})^\top $. Therefore, we can merge them directly and get the new $i$-th key and value matrices in GQA as
\begin{align}
  \tilde{W}_{K_i} &= [ W_{K_{i \times t }}, W_{K_{i \times t +1}}, \dots ,  W_{K_{i \times t + t -1}}] (\Psi^i_{d_h})^\top, \\
  \tilde{W}_{V_i} &= [ W_{V_{i \times t }}, W_{V_{i \times t +1}}, \dots ,  W_{V_{i \times t + t -1}}] (\Omega^i_{d_h})^\top,
 \label{eq:novel_MHA_2_GQA_kv}
\end{align}

where $\tilde{W}_{K_i}, \tilde{W}_{V_i} \in \mathbb{R}^{ d \times d_h}$. With the above transformation, the number of KV heads can be reduced from $h$ to $g$, so that the compression ratio of KV caches reaches $1-g/h$. Besides, according to Equations~\ref{eq:MHA} and~\ref{eq:attn}, there is also no nonlinear layer between $\Psi^i_{d_h}$ and $W_Q$, or $\Omega^i_{d_h}$ and $W_O$. So similarly, we can incorporate those low-rank weights into $W_Q$ and $W_O$. We set $p$ to be the integer part of $i/t$, $q = i-t \times p$ and $i \in \{0,1,\dots,h-1\}$, then
\begin{align}
  \tilde{W}_{Q_i} &= W_{Q_i} (\Psi^{p}_{d_h,q})^\top, \\
  \tilde{W}_{O_i} &= \Omega^p_{d_h,q} W_{O_i},
\end{align}
where $\Psi^{p}_{d_h,q}, \Omega^p_{d_h,q} \in  \mathbb{R}^{ d_h \times d_h}$ is the $q \times d_h$ through $(q+1) \times d_h$ columns of $\Psi^{p}_{d_h}$ and $\Omega^p_{d_h}$. $W_{O_i} \in  \mathbb{R}^{ d_h \times d}$ represents the $i \times d_h$ to $(i+1) \times d_h$ rows of $W_{O}$. We update the original weights $W_{Q_i}$ and $W_{O_i}$ into $\tilde{W}_{Q_i}$ and $\tilde{W}_{O_i}$, respectively. We name this method SVD-\emph{a}, as it performs SVD on the output activations.

Except for this strategy which compresses KV heads with calibration sets, another low-rank compression strategy that does not require data is to directly perform SVD on KV weights, i.e.,
\begin{align}
  \hat{W}_{K_i} &= [W_{K_{i \times t }} , W_{K_{i \times t +1}}, \dots , W_{K_{i \times t + t -1}}], \\
  \hat{W}_{V_i} &= [W_{V_{i \times t }} , W_{V_{i \times t +1}}, \dots ,  W_{V_{i \times t + t -1}}].
\end{align}
Then we perform SVD on $\hat{W}_{K_i}$ and $\hat{W}_{V_i}$, i.e., $\hat{W}_{K_i}  = \hat{\Phi}^i \hat{\Sigma}^i \hat{\Psi}^i$ and $\hat{W}_{V_i} =\hat{\Theta}^i \hat{\Lambda}^i \hat{\Omega}^i $. Similarly, we keep the top-$d_h$ rows of $\hat{\Psi}^i$ and $\hat{\Omega}^i$, and then treat them as compressed weights to replace the original $\Psi^i$ and $\Omega^i$. We treat this strategy as a baseline method and name it SVD-\emph{w}, as it low-rank decomposes model weight. Later we will show that this SVD-\emph{w} is not as effective as our approach that uses KV caches' low-rank characteristic.

After obtaining those compressed models, we use LoRA~\cite{hu2021lora} to fine-tune them. We will show that by removing half or even three-quarters of KV heads, our compressed LLMs can still achieve comparable accuracies as the original model. Note that our compression method is orthogonal to the fine-tuning strategy. If more computing resources are available, full-parameter fine-tuning is also a viable strategy to potentially achieve better accuracy.


\subsection{Deal with RoPE}
There is a special case when LLMs adopt RoPE~\cite{su2024roformer}, which inserts a relative position embedding between $W_Q$ and $W_K$. At this time, the attention score calculation becomes 
\begin{align}
  q_m k_n^\top=(x_m W_q  R_{\theta, m}^d)( x_n W_k R_{\theta, n}^d )^\top,
\end{align}
where $x_i \in  \mathbb{R}^{ 1 \times d}$ is the $i$-th ($1 \leq i \leq l -1$) token in an input sequence $x$. $R_{\theta, i}^d$ is the rotary embedding corresponding to $x_i$. $q_m$ and $k_n$ are the query and key vectors of $x_m$ and $x_n$, respectively. LLMs then view $x_n W_k R_{\theta, n}^d$ as the key cache in practice. This will, however, cause the original fusion for key cache invalid since it introduces extra position embedding between $W_k$ and $W_q$. 

To deal with this difficulty, we come up with several strategies. First, we do not divide key heads but directly calculate $\Psi$ matrices for the whole key cache in each block. This strategy avoids grouping key heads, which reduces the algorithm's constraints and can further improve the accuracy of the compressed model. Second, instead of keeping the original key cache $K$, we calculate and preserve $\tilde{K} = K (\Psi_{d_h})^\top$ during inference. Although the compressed model is no longer in the GQA pattern, key cache compression is achieved. Therefore, the original calculation becomes 
\begin{align}
  q_m k_n^\top=(x_m W_q  R_{\theta, m}^d) \Psi_{d_h}^\top \Psi_{d_h} ( x_n W_k R_{\theta, n}^d )^\top \,.
\end{align}
In particular, we will calculate $(x_m W_q  R_{\theta, m}^d)\Psi_{d_h}^\top $ instead of continuing to calculate $\Psi_{d_h}^\top\tilde{K}^\top$, because $x$ only contains one token during the generation process, but key cache often has many tokens. RoPE does not involve value caches, so the compression strategy for value caches remains the same. Later our experiments will prove that all our three strategies can reduce KV caches with faster generation.

\section{Experiment}
We now evaluate our methods in this section. More results can be found in the appendix.

\subsection{Settings}
\textbf{Foundation models.} We evaluate our framework on the 7B1 model of BLOOMZ~\cite{muennighoff2023crosslingual}, and the 7B, 13B models of LLaMA2~\cite{touvron2023llama}. BLOOMZ is BLOOM's~\cite{le2023bloom} supervised fine-tuning (SFT) version using the open-sourced xP3~\cite{muennighoff2023crosslingual} dataset.


\textbf{Compression.} For BLOOMZ-7B1, we sampled 1\textperthousand~data from the xP3 dataset. The original xP3 includes 83.6M samples, thus we randomly sampled 83.6K sampled from it. For LLaMA2 models, following QLoRA~\cite{dettmers2024qlora}, we use the FLANv2~\cite{longpre2023flan} dataset and extract 23K data from it. We concatenate the input and output of each sample for inference, and then collect KV caches in each block to calculate compression weights. 

Here we reduce half and three-quarters KV heads. The original BLOOMZ-7B1 and LLaMA2-7B have MHA structures that contain 32 heads, and we compress all attention layers in the models to 16 or 8 heads. For BLOOMZ-7B1, we continue to reduce KV heads into 4. LLaMA2-13B contains 40 heads and we reduce them to 20 or 10 heads. Note that BLOOMZ uses ALiBi~\cite{press2022train} and does not involve RoPE so it is a standard GQA mechanism after compression.

\textbf{Fine-tuning.} We apply LoRA~\cite{hu2021lora} to fine-tune the compressed model. For 7B size models, we set LoRA's $r = 256$ and $\alpha = 512$ and fine-tune 1 epoch. For 13B size models, LoRA's $r$ is 512 and $\alpha$ is 256, and we fine-tune the compressed model for 2 epochs. In all experiments, we apply a constant learning rate schedule and set the learning rate as 4e-5. We use 4$\times$80G A100 GPU to fine-tune the compressed model and set batch size per device as 1. We set the gradient accumulation steps as 16 and the LoRA dropout rate as 0.05. Weight decay is 0.05 and AdamW~\cite{loshchilov2018decoupled} is used. Note for the BLOOMZ-7B1 model, we continue to apply the same 83.6K sub-dataset to fine-tune the compressed model. For LLaMA2 models, we select 232K data from FLANv2~\cite{longpre2023flan} as the training dataset.

\textbf{Evaluation metrics.} For BLOOMZ-7B1, we follow the evaluation metrics of the original paper, i.e., we evaluate the zero-shot accuracy on XCOPA~\cite{ponti2020xcopa}, XNLI~\cite{conneau2018xnli}, XWinoGrad~\cite{tikhonov2021heads} and XStoryCloze~\cite{lin2022shot} with Bulgarian (BG), German (DE), Greek (EL), Russian(RU), Thai (TH), Turkish (TR), Japanese (JP), Estonian (ET), Haitian (HT), Italian (IT), Quechua (QU) and Burmese (MY) language questions and English prompts. For LLaMA2 models, we evaluate the perplexity on WikiText2~\cite{stephen2017pointer} and C4~\cite{raffel2020exploring}. We further assess the zero-shot commonsense question answering (QA) ability on tasks covering SIQA~\cite{sap2019social}, HellaSwag~\cite{zellers2019hellaswag}, PIQA~\cite{Bisk2020piqa}, WinoGrande~\cite{sakaguchi2021winogrande}, ARC~\cite{clark2018think}, BoolQ~\cite{clark2019boolq}, and OpenBookQA~\cite{mihaylov2018can}. We also evaluate both the zero-shot and five-shot performance of the LLMs on the Massively Multitask Language Understanding (MMLU) benchmark~\cite{hendrycks2021mmlu}. It consists of 57 language tasks including humanities, STEM, social science, etc. We adopt lm-eval-harness~\cite{gao2021framework} to produce the accuracy results. Besides, we also report throughput in an 80G A100 GPU. Since the prefilling stage is computation-bound while the decoding stage is memory-bound, and here our goal is to compress KV caches, we set the context length to 2048 and calculate throughput when generating the 2050th token for simplicity.



\subsection{Main Results}
We report the number (\#) of KV heads, throughput, and accuracies of BLOOMZ-7B1 and LLaMA2 in Tables~\ref{tab:bloomz-results} and~\ref{tab:llama-ppl-mmlu-results}, respectively. Further accuracies of eight zero-shot commonsense question answering datasets are shown in Table~\ref{tab:llama-zsqa-results}.  Note that for BLOOMZ-7B1, the `AVG.' column is not the direct average accuracy of the four datasets, as each dataset contains a different number of sub-datasets. Therefore, we report the average accuracy of those sub-datasets. Detailed results are shown in the appendix. For LLaMA2 models, we abbreviate WikiText2, HellaSwag, WinoGrande, and OpenBookQA to W2, HLSW, WG, and OBQA, respectively. ARC-e and ARC-c stand for ARC-easy and ARC-challenge tasks, respectively. 

\begin{table}
	\centering
	\caption{Performances of BLOOMZ-7B1 with our KV heads compression framework.}
	\setlength{ \tabcolsep}{1.2mm}
	\small
	\begin{tabular}{c|c|cccc|c}
	  \hline
	  \multirow{2}{*}{\textbf{\#KV Heads}}   &  \multirow{2}{*}{\textbf{Throughput} (token/s)} & \multicolumn{4}{c|}{\textbf{Accuracy}} & \multirow{2}{*}{\textbf{AVG.}} \\ \cline{3-6}
	   &  & \textbf{XNLI} & \textbf{XWinoGrad}  & \textbf{XCOPA} & \textbf{XStoryCloze} &  \\ \hline
	   32 & 8.56 &39.73 & 51.49 & 51.03 & 54.25 & 47.25 \\ \hline
	   16 &  14.41 & 39.48 & 51.47 & 52.97 & 54.06 & \textbf{47.85}\\ 
	   8 & 23.24 & 39.03 & 50.45 & 51.60 & 52.37 & 46.84\\
	   4 & \textbf{34.78} & 38.45 & 50.29 & 50.27 & 50.89 & 45.92\\ \hline
	\end{tabular}
	\label{tab:bloomz-results}
\end{table}  

\begin{table}
	\centering
	\caption{Performances of LLaMA2 models with our KV heads compression framework.}
	\setlength{ \tabcolsep}{1.2mm}
	\small
	\begin{tabular}{c|c|c|cc|cc}
	  \hline
	  \multirow{2}{*}{\textbf{Model}} &  \multirow{2}{*}{\textbf{\#KV Heads}} &  \multirow{2}{*}{\textbf{Throughput} (token/s)} & \multicolumn{2}{c|}{\textbf{Perplexity} ($\downarrow$)} & \multicolumn{2}{c}{\textbf{MMLU} }  \\ \cline{4-7}
	   &  & & \textbf{W2} & \textbf{C4}  & \textbf{0-shot} & \textbf{5-shot}  \\ \hline
	   \multirow{3}{*}{LLaMA2-7B} & 32 & 8.05 & \textbf{5.47} & \textbf{6.98}  & 41.79 & 45.82\\  \cline{2-7}
		&16 &  13.41 & 7.08& 9.12 & \textbf{48.32} & \textbf{48.74}\\ 
	   & 8 &  \textbf{20.81} & 9.17 &11.24 & 44.62 & 45.54\\\hline
  
	   \multirow{3}{*}{LLaMA2-13B} & 40 &  5.04 & \textbf{4.88} & \textbf{6.47} & 52.12 & \textbf{55.17} \\ \cline{2-7}
		&20 &  8.60  & 6.51 & 8.01 & \textbf{53.65} & 54.71\\ 
	   & 10 & \textbf{13.93} & 8.40 & 9.86 & 49.65 & 50.73 \\\hline
	\end{tabular}
	\label{tab:llama-ppl-mmlu-results}
\end{table}

\begin{table}
  \centering
  \caption{Accuracies in the commonsense QA datasets with different \#KV heads on LLaMA2 models.}
  \setlength{ \tabcolsep}{1.2mm}
  \small
  \begin{tabular}{c|c|ccccccccc}
    \hline
    \textbf{Model} &  \textbf{\#KV Heads} &  \textbf{BoolQ} & \textbf{PIQA} & \textbf{SIQA} & \textbf{HLSW} & \textbf{WG} &\textbf{ARC-e} & \textbf{ARC-c} & \textbf{OBQA} & \textbf{Avg.} \\ \hline
     \multirow{3}{*}{LLaMA2-7B} & 32 & 77.77  & 79.05 & 32.91 & 76.00  &  69.22& 74.58 & 46.25 &  44.20 & 62.50 \\ \cline{2-11}
      &16 & 83.61 & 77.97 & 32.80 & 72.53 & 73.56 & 78.45 & 49.66 & 46.40 & \textbf{64.37}\\ 
     & 8 & 80.58&76.77 & 32.91 & 66.67 & 67.88 & 73.32 & 43.34 & 43.80 & 60.66 \\\hline

     \multirow{3}{*}{LLaMA2-13B} & 40 & 80.61&80.52&33.11&79.38&72.30&77.40&49.06&45.20&64.70  \\ \cline{2-11}
      &20 & 85.78 & 80.41 & 32.04 & 77.84 & 74.59 & 80.22 & 54.35 & 46.80 &\textbf{66.50}\\ 
     & 10 & 83.24 & 78.45 & 33.11 & 73.00 & 70.80 & 75.97 & 49.06 & 41.80 & 63.18\\\hline
    \end{tabular}
  \label{tab:llama-zsqa-results}
\end{table}  

As these results exhibited, when compressing half of KV heads, our algorithm can maintain the same or even higher accuracy, while guaranteeing more than 50\% throughput improvement at the decoding process. This phenomenon indicates that our framework is a powerful solution in scenarios where memory efficiency is required. When further compressing three-quarters of KV heads, BLOOM-7B1 and LLaMA2-13B lost more KV cache information, resulting in a slight accuracy drop. However, those results are still comparable with the original models' accuracies. These results demonstrate that our approach is an effective strategy to compress KV heads and reduce KV cache sizes, thereby alleviating the heavy memory bandwidth pressure during the LLM generation phase.


\subsection{Ablation Studies}

We further perform several analyses to explore the impact of different modules of our method.

\begin{table}
  \centering
  \caption{Performances of LLaMA2 models with different initialization strategies.}
  \small
  \begin{tabular}{c|cc|cc|cc|ccc}
    \hline
    \multirow{2}{*}{\textbf{\#KV Heads}} &\multirow{2}{*}{\textbf{Methods}} & \multirow{2}{*}{\textbf{Stage}} & \multicolumn{2}{c|}{\textbf{Perplexity} ($\downarrow$)} & \multicolumn{2}{c|}{\textbf{MMLU} } &  \multirow{2}{*}{\textbf{QA AVG.}} \\ \cline{4-7}
    &    & & \textbf{W2} & \textbf{C4}  & \textbf{0-shot} & \textbf{5-shot}&  \\ \hline
       \multirow{6}{*}{16} &  \multirow{2}{*}{Mean-pool} & Initialization & 1079.43 & 625.68 & 22.94 & 24.94 & 34.88\\ 
       &  & Fine-tune & 1079.43 & 625.68&35.25 & 35.41 & 56.57 \\ \cline{2-9}
       &  \multirow{2}{*}{SVD-\emph{w}} & Initialization & 807.62& 624.78& 23.11 & 23.17 & 35.46\\ 
       & & Fine-tune& 10.44 & 12.28 & 42.46& 39.38 & 62.10\\ \cline{2-9}
       &  \multirow{2}{*}{SVD-\emph{a}} & Initialization &  \textbf{13.57} &\textbf{18.57} &\textbf{28.05} & \textbf{25.97} & \textbf{48.87}\\ 
       & & Fine-tune &  \textbf{7.08} & \textbf{9.12} & \textbf{48.32} & \textbf{48.74} & \textbf{64.37}\\ \hline
       \multirow{6}{*}{8} &  \multirow{2}{*}{Mean-pool} & Initialization & 4113.79 & 2381.27 & 25.00 & 24.95 & 34.69\\ 
       &  & Fine-tune & 63.32 & 37.87 &26.63 & 25.30 & 40.79\\ \cline{2-9}
       &  \multirow{2}{*}{SVD-\emph{w}} & Initialization & 3888.63 & 2350.09& 23.42 & 22.94 & 35.43 \\ 
       & & Fine-tune & 15.44 & 17.13& 24.30 & 23.69 & 55.86 \\ \cline{2-9}
       &  \multirow{2}{*}{SVD-\emph{a}} & Initialization & \textbf{194.61} & \textbf{141.84} & \textbf{23.61} & \textbf{23.93} &\textbf{37.59} \\ 
       & & Fine-tune & \textbf{9.17} & \textbf{11.24} & \textbf{44.62} & \textbf{45.54} & \textbf{60.66} \\ \hline
    \end{tabular}
  \label{tab:ablation-llama-initialization-results}
\end{table}  

\begin{table}
	\centering
	\caption{Performances of LLaMA2 models with different initialization data sizes.}
	\small
	\begin{tabular}{c|c|cc|cc|c}
	  \hline
	  \multirow{2}{*}{\textbf{\#KV Heads}} &  \multirow{2}{*}{\textbf{Data}}  & \multicolumn{2}{c|}{\textbf{Perplexity} ($\downarrow$)} & \multicolumn{2}{c|}{\textbf{MMLU} } &  \multirow{2}{*}{\textbf{QA AVG.}} \\ \cline{3-6}
	  &  &  \textbf{W2} & \textbf{C4}  & \textbf{0-shot} & \textbf{5-shot}&  \\ \hline
		 \multirow{3}{*}{16} & 23K & \textbf{13.57} & \textbf{18.57} &28.05 & 25.97 & \textbf{48.87}\\ 
		 & 46K& 13.83& 18.89 &27.96 & 25.89 & 48.50\\ 
		 & 92K&  13.93& 18.96& \textbf{28.11} & \textbf{25.98} & 48.66\\
		 & 232K&  13.92& 18.93& 27.89 & 25.87 & 48.70\\ \hline
		 \multirow{4}{*}{8} &  23K & \textbf{194.61} & \textbf{141.84} & 23.61& 23.93  & \textbf{37.59}\\
		 & 46K& 197.23 & 142.80 & \textbf{23.76} &  23.81& 37.46\\ 
		 & 92K& 202.05 & 145.44&23.62&23.79  & 37.34\\ 
		 & 232K& 202.38 & 145.43& 23.74 & \textbf{24.01} & 37.48\\ \hline
		\end{tabular}
	\label{tab:ablation-llama-initialization-datasize}
\end{table}

\begin{table}
	\centering
	\caption{Performances of LLaMA2 models with different fine-tuning data sizes.}
	\small
	\begin{tabular}{c|c|cc|cc|c}
	  \hline
	  \multirow{2}{*}{\textbf{\#KV Heads}} &  \multirow{2}{*}{\textbf{Data}}  & \multicolumn{2}{c|}{\textbf{Perplexity} ($\downarrow$)} & \multicolumn{2}{c|}{\textbf{MMLU} } &  \multirow{2}{*}{\textbf{QA AVG.}} \\ \cline{3-6}
	  &  &  \textbf{W2} & \textbf{C4}  & \textbf{0-shot} & \textbf{5-shot}&  \\ \hline
		 \multirow{4}{*}{16} & 23K & 7.29 & 9.41 & 43.49 &44.16 & 61.93\\ 
		 & 46K& 7.38 & 9.35 & 42.42 & 44.62 & 61.53\\ 
		 & 92K& 7.26  & 9.20 &45.00 & 46.36 & 62.57\\ 
		 & 232K& \textbf{7.08} & \textbf{9.12} & \textbf{48.32} & \textbf{48.74} & \textbf{64.37}\\ \hline
		 \multirow{4}{*}{8} &  23K & 9.62& 12.11&32.71 & 35.37 & 57.13\\
		 & 46K&9.48&11.84 & 39.93& 40.10& 57.53\\ 
		 & 92K&  9.73 &11.50 & 40.53&42.90 & 59.04\\ 
		 & 232K& \textbf{9.17} & \textbf{11.24} & \textbf{44.62} & \textbf{45.54} & \textbf{60.66}\\ \hline
	  \end{tabular}
	\label{tab:ablation-llama-fine-tuning-datasize}
\end{table}

\textbf{Comparing different compression strategies.} Here we compare three different compression strategies, namely directly mean-pooling KV head weights~\cite{ainslie2023gqa}, performing SVD for KV head weights (i.e., SVD-\emph{w}), and our methods (i.e., SVD-\emph{a}). All other settings are consistent except for the compression strategy. We apply LLaMA2-7B with 16 and 8 KV heads and report the results of direct compression (i.e., the `Initialization' rows) and further fine-tuning (i.e., the `Fine-tune' rows). Due to page limitation, we report the average accuracy of eight zero-shot commonsense QA datasets here (i.e., the `QA AVG.' column).

As we can see from Table \ref{tab:ablation-llama-initialization-results}, both mean-pooling and SVD-\emph{w} lead to collapse before fine-tuning, and our method far exceeds them both before \& after fine-tuning. Those results reveal the effectiveness of our algorithm. That is, with limited computing and data resources, our SVD-\emph{a} can significantly reduce KV cache sizes and are very practical, while other strategies are not.

\textbf{Influence of different data sizes for initialization and fine-tuning.} Here we investigate the effect of the size of different data sets. Similarly, we apply LLaMA2-7B with 16 and 8 KV heads as the baseline. In particular, in our original settings, we calculate the compression weights with 23K samples for initialization and fine-tune the compression model with 232K samples. Here we sample 46K and 92K data from the original FLANv2 dataset and calculate the compression weights. Then based on the model compressing with 23K samples, we continue to fine-tune the model with 23K, 46K, 92K, and 232K samples respectively.  

Results of different dataset sizes on model initialization and fine-tuning are shown in Tables~\ref{tab:ablation-llama-initialization-datasize} and~\ref{tab:ablation-llama-fine-tuning-datasize}, respectively. As we can see, the data size required for initialization is not high, while the fine-tuning process is relatively more data-hungry. During initialization, excess data does not necessarily yield additional benefits. However, the more data for fine-tuning, the better the compression model will perform. This phenomenon reveals that our method can further achieve higher precision if there are more data and computational resources.

\begin{table}
  \centering
  \caption{LLaMA2: different epoch v.s. data.}
  \small
  \begin{tabular}{c|cc|cc|cc|c}
    \hline
    \multirow{2}{*}{\textbf{\#KV Heads}} &  \multirow{2}{*}{\textbf{Data}} & \multirow{2}{*}{\textbf{Epoch}}  & \multicolumn{2}{c|}{\textbf{Perplexity} ($\downarrow$)} & \multicolumn{2}{c|}{\textbf{MMLU} } &  \multirow{2}{*}{\textbf{QA AVG.}} \\ \cline{4-7}
    &  & & \textbf{W2} & \textbf{C4}  & \textbf{0-shot} & \textbf{5-shot}&  \\ \hline
       \multirow{3}{*}{16} & 46K & 1 &7.38 & 9.35 & 42.42 & 44.62 & 61.53 \\ 
       & 46K & 2 & 7.64 & 9.60& \textbf{45.37} & 45.77 & \textbf{62.90}\\ 
       & 92K & 1 & \textbf{7.26}  & \textbf{9.20} &45.00 & \textbf{46.36} & 62.57 \\ \hline
       \multirow{3}{*}{8} &  46K & 1  & 9.48&11.84 & 39.93& 40.10& 57.53\\ 
       &  46K & 2 & \textbf{9.56} &12.72 & \textbf{41.57}& 41.77 & 58.76\\ 
       &   92K & 1 & 9.73&\textbf{11.50}&40.53&\textbf{42.90}&\textbf{59.04}\\ \hline
    \end{tabular}
  \label{tab:ablation-llama-epoch-datasize}
\end{table}

\textbf{Enlarge data sizes v.s. increase training epochs.} Here we investigate the effect of training time on the results, i.e., given the same training GPU resources, whether it is more beneficial to continue enlarging the training dataset size or increasing the training epoch. Similarly, we train the compressed LLaMA2-7B with 2 epoch and 46K training samples, and compare the results of training 1 epoch with 92K data. The results are shown in Table~\ref{tab:ablation-llama-epoch-datasize}. It can be seen that the addition of high-quality samples is more helpful to the model accuracies than increasing training epochs.

\section{Conclusion, Limitation, and Future Work}
In this paper, we proposed a low-rank decomposition framework to compress KV heads. We first discovered that KV caches are low-rank, and based on this finding we converted the original MHA architecture to GQA mechanism by low-rank decomposition. We also proposed several special strategies to handle the attention layer with RoPE. With half or even three-quarters of KV cache compressed, our approach can recover LLM's accuracy with limited training and data resources, while this is not possible with the previous KV head compression method.

Although our compression framework can significantly reduce KV caches and improve the speed of LLM generation, our approach currently has some shortcomings. First of all, when our method is used in the attention layer with RoPE, the key head compression weights cannot be incorporated into the original model, resulting in additional parameters and computational overhead, which will slightly slow down the prefill stage. So how to compress key heads with RoPE more elegantly will be an interesting direction in the future. In addition, when heavily compressing KV heads or the model size is too large, LLM's accuracy will decrease to some extent. Therefore, how to better deal with these situations is also a future direction. Besides, our approach can potentially be combined with previous KV cache compression methods to achieve an extreme KV cache compression ratio, which we leave as a viable direction for the future.

{\small
\bibliographystyle{plain}
\bibliography{refs}
}

\clearpage
\appendix
\section{The running time of compression}

Here we report the time required to compress LLaMA2 models. The results are shown in Table~\ref{tab:appendix-time}. Numbers in the table are measured in hours. As we can see, our algorithm requires very little time to compute the compression weights, and the bottleneck of the entire framework is the time spent fine-tuning the compressed models.

\begin{table}[h]
  \centering
  \caption{Hours needed to calculate compressing weights and fine-tune the model.}
  \setlength{\tabcolsep}{0.8mm}
  \small
  \renewcommand{\arraystretch}{1.0} 
  \begin{tabular}{l|c|c|c|c|l|c|c|c|c}
    \hline
    \multirow{2}{*}{\textbf{Size}} & \multicolumn{2}{c|}{\textbf{Initialization}} & \multicolumn{2}{c|}{\textbf{Fine-tuning}} & \multirow{2}{*}{\textbf{Size}} & \multicolumn{2}{c|}{\textbf{Initialization}} & \multicolumn{2}{c}{\textbf{Fine-tuning}} \\ \cline{2-5}\cline{7-10}
    &   Head = 16 & Head = 8 & Head = 16 &  Head = 8 & & Head = 20 & Head = 10 & Head = 20 & Head = 10 \\ \hline
    7B & \multicolumn{2}{c|}{0.30} & 43.40& 36.14 & 13B &  \multicolumn{2}{c|}{0.68} & 103.47  & 87.01\\\hline
  \end{tabular}
  \label{tab:appendix-time}
\end{table}  

\section{Detailed results of compressed models}

In Table~\ref{tab:appendix-bloom-detailed-results} we report detailed accuracies of BLOOMZ-7B1 and its compressed version on each sub-dataset. For LLaMA2 models in the MMLU datasets, we report the accuracies in Table~\ref{tab:appendix-llama2-mmlu-detailed-results}.

\begin{table}[h]
    \centering
    \caption{Detailed accuracies of BLOOMZ-7B1 with different KV heads.}
    \small
    \renewcommand{\arraystretch}{1.0} 
    \begin{tabular}{c|cccccc|cc|c}
      \hline
       \textbf{\#KV Heads} & \multicolumn{8}{c|}{\textbf{Dataset}} & \textbf{Average ACC.}  \\ \hline
       \multirow{6}{*}{32}& \multicolumn{6}{c|}{\textbf{XNLI}} & \multicolumn{2}{c|}{\textbf{XWinoGrad}} & \multirow{6}{*}{47.25}  \\ \cline{2-9}

       &   \textbf{BG} & \textbf{DE} & \textbf{EL} & \textbf{RU} & \textbf{TH} &\textbf{TR} & \textbf{JP} & \textbf{RU}\\ \cline{2-9}
      & 40.30 & 42.80 & 38.74 & 42.70 & 37.69 & 36.15 &  50.26  & 52.70 &\\\cline{2-9}
      & \multicolumn{6}{c|}{\textbf{XCOPA}} & \multicolumn{2}{c|}{\textbf{XStoryCloze}}\\\cline{2-9}
      & \textbf{ET} & \textbf{HT} & \textbf{IT} & \textbf{QU} & \textbf{TH} &\textbf{TR} & \textbf{MY} & \textbf{RU}\\\cline{2-9}
      & 48.20& 49.40 & 57.00 & 48.00 & 55.20 & 48.40 & 49.46 & 59.03\\\hline

      \multirow{6}{*}{16}& \multicolumn{6}{c|}{\textbf{XNLI}} & \multicolumn{2}{c|}{\textbf{XWinoGrad}} &   \multirow{6}{*}{\textbf{47.85}}  \\ \cline{2-9}
      &   \textbf{BG} & \textbf{DE} & \textbf{EL} & \textbf{RU} & \textbf{TH} &\textbf{TR} & \textbf{JP} & \textbf{RU}\\ \cline{2-9}
     & 39.86 & 42.92 & 38.51 & 42.62 & 36.99 & 36.00 & 50.43 & 52.51 & \\\cline{2-9}
      & \multicolumn{6}{c|}{\textbf{XCOPA}} & \multicolumn{2}{c|}{\textbf{XStoryCloze}}\\\cline{2-9}
      & \textbf{ET} & \textbf{HT} & \textbf{IT} & \textbf{QU} & \textbf{TH} &\textbf{TR} & \textbf{MY} & \textbf{RU}\\\cline{2-9}
      & 50.00& 51.40 & 57.20 & 51.40 & 59.20 & 48.60 & 50.09 & 58.03\\\hline

      \multirow{6}{*}{8}& \multicolumn{6}{c|}{\textbf{XNLI}} & \multicolumn{2}{c|}{\textbf{XWinoGrad}} &   \multirow{6}{*}{46.84}  \\ \cline{2-9}
      &   \textbf{BG} & \textbf{DE} & \textbf{EL} & \textbf{RU} & \textbf{TH} &\textbf{TR} & \textbf{JP} & \textbf{RU}\\ \cline{2-9}
     & 39.27 & 42.16 & 38.35 & 41.57 & 37.00 & 35.85 & 49.16 & 51.75 & \\\cline{2-9}
      & \multicolumn{6}{c|}{\textbf{XCOPA}} & \multicolumn{2}{c|}{\textbf{XStoryCloze}}\\\cline{2-9}
      & \textbf{ET} & \textbf{HT} & \textbf{IT} & \textbf{QU} & \textbf{TH} &\textbf{TR} & \textbf{MY} & \textbf{RU}\\\cline{2-9}
      & 50.00& 49.80 & 55.60 & 51.40 & 54.40 & 48.40 & 49.34 & 55.39\\\hline

      \multirow{6}{*}{4}& \multicolumn{6}{c|}{\textbf{XNLI}} & \multicolumn{2}{c|}{\textbf{XWinoGrad}} &   \multirow{6}{*}{45.92}  \\ \cline{2-9}
      &   \textbf{BG} & \textbf{DE} & \textbf{EL} & \textbf{RU} & \textbf{TH} &\textbf{TR} & \textbf{JP} & \textbf{RU}\\ \cline{2-9}
     & 38.82 & 40.76 & 37.70 & 41.43 & 36.38 & 35.61 & 48.91 & 51.68 & \\\cline{2-9}
      & \multicolumn{6}{c|}{\textbf{XCOPA}} & \multicolumn{2}{c|}{\textbf{XStoryCloze}}\\\cline{2-9}
      & \textbf{ET} & \textbf{HT} & \textbf{IT} & \textbf{QU} & \textbf{TH} &\textbf{TR} & \textbf{MY} & \textbf{RU}\\\cline{2-9}
      & 49.20& 47.40 & 48.80 & 51.20 & 55.60 & 49.40 & 47.85 & 53.94\\\hline

    \end{tabular}
    \label{tab:appendix-bloom-detailed-results}
  \end{table}

  \begin{table}[h]
    \centering
    \caption{Detailed accuracies of LLaMA2 models with different KV heads in MMLU datasets.}
    \setlength{\tabcolsep}{0.8mm}
    \small
    \renewcommand{\arraystretch}{1.0} 
    \begin{tabular}{l|c|ccccc|ccccc}
      \hline
      \multirow{2}{*}{\textbf{Model}} & \multirow{2}{*}{\textbf{\#KV Heads}} & \multicolumn{5}{c|}{\textbf{MMLU} ($0$-shot)} & \multicolumn{5}{c}{\textbf{MMLU} ($5$-shot)} \\ \cline{3-12}
      &  &  \textbf{Hums.} & \textbf{STEM} & \textbf{Social} & \textbf{Other} & \textbf{Avg.} &\textbf{Hums.} & \textbf{STEM} & \textbf{Social} & \textbf{Other} & \textbf{Avg.} \\ \hline
      \multirow{3}{*}{LLaMA2-7B} & 32 & 39.64  & 34.25 & 47.35 & 47.18  &  41.79 & 43.32 & 36.98 & 51.77  & 52.69  & 45.82  \\\cline{2-12}
      & 16 & 44.99 & 39.42 & 55.96 & 54.81 &  48.32 & 44.48 & 39.93 & 56.68& 56.26 & 48.74 \\
      & 8 & 41.06 &37.36 &52.58 & 49.50 & 44.62 & 41.11 & 38.31 & 54.01 &51.21 & 45.54 \\\hline
      \multirow{3}{*}{LLaMA2-13B} & 40 & 47.99  & 42.21 & 61.23 & 59.41  & 52.12  & 53.43 & 43.84  & 63.21  & 61.35 & 55.17  \\\cline{2-12}
      & 20 & 49.54&43.48 &62.98 & 60.93& 53.65 & 50.97 & 44.24 & 63.99 & 61.83 & 54.71\\
      & 10 & 46.25&40.22 & 57.62&56.49 &49.65 & 46.89& 41.42&59.90 &56.90 & 50.73\\\hline
    \end{tabular}
    \label{tab:appendix-llama2-mmlu-detailed-results}
  \end{table}  

  \section{Detailed results of different compression strategies}

  In Tables~\ref{tab:appendix-llama-initialization-mmlu-results} and~\ref{tab:appendix-llama-initialization-zsqa-results}, we compare the influence of three different compression strategies in detail, which can better reflect the advantage of our compression strategy. Although the model accuracy collapses after direct compression, our compression model can quickly recover the accuracy after further fine-tuning. These results demonstrate the effectiveness of our framework.

  \begin{table}[h]
    \centering
    \caption{Detailed accuracies in MMLU of LLaMA2 models with different initialization strategies.}
    \setlength{ \tabcolsep}{0.6mm}
    \small
    \begin{tabular}{c|cc|ccccc|ccccc}
      \hline
      \multirow{2}{*}{\textbf{\#KV Heads}} & \multirow{2}{*}{\textbf{Methods}} & \multirow{2}{*}{\textbf{Stage}}  & \multicolumn{5}{c|}{\textbf{MMLU} (0-shot)} &  \multicolumn{5}{c}{\textbf{MMLU} (5-shot)} \\ \cline{4-13}
      &    & &  \textbf{Hums.} & \textbf{STEM} & \textbf{Social} & \textbf{Other} & \textbf{Avg.} &\textbf{Hums.} & \textbf{STEM} & \textbf{Social} & \textbf{Other} & \textbf{Avg.}   \\ \hline
         \multirow{6}{*}{16} &  \multirow{2}{*}{Mean-pool} & Init. & 24.21 & 21.22 & 21.71 & 23.98 & 22.94 & 24.21 & 21.22 & 21.71 & 23.98 & 22.94 \\ 
         &  & Fine-tune &32.99 & 31.18 &37.60 & 40.49&35.25 &33.24 & 20.83 &38.25& 40.52& 35.41\\ \cline{2-13}
         &  \multirow{2}{*}{SVD-\emph{w}} & Init. & 24.63&21.44 &21.94 & 23.66 &23.11 &24.21 &21.73 &21.84 &24.40 &23.17\\ 
         & & Fine-tune&38.98 & 34.09& 49.17& 49.57& 42.46&37.98 & 30.76&44.59 & 45.09& 39.38\\ \cline{2-13}
         &  \multirow{2}{*}{SVD-\emph{a}} & Init. & 25.06& 29.72 & 31.04 &27.94 & \textbf{28.05} & 25.48 & 24.10 & 24.44 &  30.13& \textbf{25.97} \\ 
         & & Fine-tune & 44.99 & 39.42 & 55.96 & 54.81 &  \textbf{48.32} & 44.48 & 39.93 & 56.68& 56.26 & \textbf{48.74}\\ \hline
         \multirow{6}{*}{8} &  \multirow{2}{*}{Mean-pool} & Init. &24.82 & 24.52 & 24.80 & 25.97& \textbf{25.00} & 24.23 & 27.40 & 23.56 & 24.23& \textbf{24.95}  \\ 
         &  & Fine-tune & 23.80&29.24 & 30.26& 23.80 & 26.63 & 24.12 &  29.08 & 25.22 & 23.30 &  25.30  \\ \cline{2-13}
         &  \multirow{2}{*}{SVD-\emph{w}} & Init. & 24.00 & 22.49 &22.62 & 24.27 & 23.42 & 24.40 & 21.38& 21.61 & 23.62 & 22.94 \\ 
         & & Fine-tune & 24.97 &23.28 & 23.14  & 25.46 & 24.30& 24.25& 23.88 & 22.07 &24.24 & 23.69 \\ \cline{2-13}
         &  \multirow{2}{*}{SVD-\emph{a}} & Init.  & 24.08 & 22.90 &  23.53& 23.69& 23.61& 24.08 & 24.55 & 22.91 & 24.07 & 23.93\\ 
         & & Fine-tune & 41.06 &37.36 &52.58 & 49.50 & \textbf{44.62} & 41.11 & 38.31 & 54.01 &51.21 & \textbf{45.54} \\ \hline
      \end{tabular}
    \label{tab:appendix-llama-initialization-mmlu-results}
  \end{table}

  \begin{table}[h]
    \centering
    \caption{Detailed accuracies in zero-shot commonsense QA datasets of LLaMA2 models with different initialization strategies.}
    \setlength{ \tabcolsep}{0.6mm}
    \small
    \begin{tabular}{c|cc|cccccccc|c}
      \hline
      \multirow{2}{*}{\textbf{\#KV Heads}} &  \multirow{2}{*}{\textbf{Methods}} & \multirow{2}{*}{\textbf{Stage}}   &  \multicolumn{8}{c|}{\textbf{Commonsense QA} (0-shot)}  &  \multirow{2}{*}{\textbf{AVG.}}\\ \cline{4-11}
      &   & & \textbf{BoolQ} & \textbf{PIQA} & \textbf{SIQA} & \textbf{HLSW} & \textbf{WG} &\textbf{ARC-e} & \textbf{ARC-c} & \textbf{OBQA} &  \\ \hline
         \multirow{6}{*}{16} &  \multirow{2}{*}{Mean-pool} & Init. & 37.80  &50.76 & 32.70 & 26.47 & 50.28 & 26.68&26.71 & 27.60 & 34.88  \\ 
         &  & Fine-tune  & 73.27&74.37  &  34.29& 60.44& 62.51& 70.12 &40.36  &37.20 & 56.57 \\ \cline{2-12}
         &  \multirow{2}{*}{SVD-\emph{w}} &  Init. & 38.47 & 53.10  & 31.58 & 27.88 & 51.62& 28.83 &24.23  & 28.00 & 35.46\\ 
         & & Fine-tune& 79.45& 77.80& 32.96 & 69.20 &69.69 &76.30 & 46.59 & 44.80 & 62.10\\ \cline{2-12}
         &  \multirow{2}{*}{SVD-\emph{a}} & Init. & 60.21  & 69.26& 33.27&  51.93&  59.67& 51.60 & 31.83 &33.20 & \textbf{48.87}\\ 
         & & Fine-tune & 83.61 & 77.97 & 32.80 & 72.53 & 73.56 & 78.45 & 49.66 & 46.40 & \textbf{64.37}\\ \hline
         \multirow{6}{*}{8} &  \multirow{2}{*}{Mean-pool} & Init. & 37.83 & 49.89& 33.06 & 26.28 & 48.93& 26.22& 27.30 & 28.00 & 34.69\\ 
         &  & Fine-tune & 60.03 & 58.38 & 32.91 & 38.48 & 50.75 & 37.92& 24.49 &  23.40 & 40.79 \\ \cline{2-12}
         &  \multirow{2}{*}{SVD-\emph{w}} & Init. & 42.60 &50.76 &34.14  &26.59  & 51.22 & 26.39& 25.94 & 25.80 &  35.43\\ 
         & & Fine-tune &  69.60&73.99 & 32.75 & 60.30 &63.93 & 70.24&  39.68& 36.40 & 55.86 \\ \cline{2-12}
         &  \multirow{2}{*}{SVD-\emph{a}} & Init.  & 46.33 & 56.96 & 32.80 & 30.30& 51.70& 32.53 &24.32  & 25.80 & \textbf{37.59}\\ 
         & & Fine-tune  &  80.58&76.77 & 32.91 & 66.67 & 67.88 & 73.32 & 43.34 & 43.80 & \textbf{60.66} \\ \hline
      \end{tabular}
    \label{tab:appendix-llama-initialization-zsqa-results}
  \end{table}  

\end{document}